\documentclass[12pt,english]{article}
\usepackage[left=1in, right=1in]{geometry}
\usepackage{hyperref}
\hypersetup{colorlinks=true,linkcolor=blue,citecolor=blue,urlcolor=blue}
\usepackage[square]{natbib}
\usepackage{graphicx}%
\usepackage{multirow}%
\usepackage{amsmath,amssymb,amsfonts}%
\usepackage{amsthm}%
\usepackage{mathrsfs}%
\usepackage[title]{appendix}%
\usepackage{xcolor}%
\usepackage{textcomp}%
\usepackage{manyfoot}%
\usepackage{booktabs}%
\usepackage{algorithm}%
\usepackage{algorithmicx}%
\usepackage{algpseudocode}%
\usepackage{listings}%
\usepackage{subcaption}

\newtheorem{definition}{Definition}%

\newcommand{\vect}{\text{vec}}
\newcommand{\tensor}{\text{tensor}}
\DeclareMathOperator*{\argmax}{arg\,max}

\newtheorem{lemma}{Lemma}

\title{A Priori Generalizability Estimate for a CNN}
\author{Cito Balsells$^{1,2}$ \and Beatrice Riviere$^1$ \and David Fuentes$^2$}
\date{%
    \small $^1$ Department of Computational Applied Mathematics and Operations Research, The Ken Kennedy Institute, Rice University, 6100 Main St, Houston, TX, 77005, USA\\%
    \small $^2$ Department of Imaging Physics, The University of Texas MD Anderson Cancer Center, 1515 Holcombe Blvd, Houston, TX, 77030, USA}

\begin{document}

\maketitle
\begin{abstract}
    \footnotesize We formulate truncated singular value decompositions of entire convolutional neural networks. We demonstrate the computed left and right singular vectors are useful in identifying which images the convolutional neural network is likely to perform poorly on. To create this diagnostic tool, we define two metrics: the Right Projection Ratio and the Left Projection Ratio. The Right (Left) Projection Ratio evaluates the fidelity of the projection of an image (label) onto the computed right (left) singular vectors. 
We observe that both ratios are able to identify the presence of class imbalance for an image classification problem. Additionally, the Right Projection Ratio, which only requires unlabeled data, is found to be correlated to the model's performance when applied to image segmentation. This suggests the Right Projection Ratio could be a useful metric to estimate how likely the model is to perform well on a sample.
\end{abstract}

\section*{Introduction}\label{sec:introduction}
\normalsize
There has been a growing trend of increasing the compute power available to machine learning models in an effort to improve model performance. This can be seen with estimates that the total compute power of all NVIDIA chips has doubled every 10 months since 2019 \cite{epoch_gpu_calculation}. However, this approach is generally only accessible to a handful of large technology companies. It is not practical, nor potentially even feasible, for every organization to increase their compute power. For resource constrained settings, it is therefore crucial to find ways to maximize model performance with respect to the available resources.
There are two popular ways to accomplish this: parameter reduction and identifying a coreset for training.

While our research is more applicable to finding a coreset, we discuss parameter reduction since our methods use topics related to it. We will analyze a specific class of machine learning model referred to as a convolutional neural network (CNN) that is commonly used for image classification and image segmentation.
A common approach to parameter reduction is to compute low rank approximations for \textit{specific weight tensors} in a model to reduce memory usage. This has been shown to effectively reduce memory usage and speed up model inference while maintaining comparable performance \cite{exploiting_linear_structure,low_rank_filter_bank}. In \cite{exploiting_linear_structure}, the authors demonstrate a $2\times$ speedup within the convolutional layers by replacing each convolution kernel with a low rank approximation obtained by a higher-order singular value decomposition (SVD).

Our contribution is in computing the truncated SVD for \textit{the entire CNN} and demonstrating that it can serve as a useful diagnostic tool. To our knowledge, computing the SVD of a whole CNN is a novel application. We will show that this truncated SVD can identify which unlabeled images the model will likely perform well or poorly on. This is accomplished by evaluating the fidelity of the image and its projection onto the right singular vectors. This can then be used to assess whether or not there is bias in the model.
There are existing diagnostic tools that assess models; however, they evaluate models component by component. For example, WeightWatcher from \cite{weight_watcher_paper} performs a number of tasks including quantifying how well or poorly trained a layer is, finding outliers in the training set, and making test set predictions with or without training or testing data. This is accomplished by looking at \textit{individual weight matrices} in dense neural networks and analyzing their spectrum of eigenvalues.

In general settings, it is only possible to compute the truncated SVD of a CNN with implicit methods. This requires an operator that performs the transpose or adjoint of the CNN. This operator, which we refer to as the Adjoint CNN, is defined and constructed in Section \ref{methods:theory}. In Section \ref{methods:implementation_solver} we detail how we compute the SVD and how it is used to evaluate the fidelity of projections.
Then in Section \ref{methods:mnist} and Section \ref{methods:brats} we present the methods behind our experiments on two different datasets.
Following the methods, we present our results in Section \ref{sec:results} with our interpretation of the results in Section \ref{sec:discussion}. 

Primary contributions of this paper:
\begin{enumerate}
    \item Theoretical and empirical proof that the adjoint of the matrix associated with a CNN can be interpreted as a closely related CNN,
    \item Demonstration on two datasets that metrics derived from the SVD, the Right and Left Projection Ratios, can identify when a CNN is likely to perform well or poor.
\end{enumerate}

\section{Methods}\label{sec:methods}

We provide a general overview for how we compute the SVD of an arbitrary CNN. The CNNs we study are compositions of: convolution (with no bias), downsampling, upsampling, ReLU activation, and skip connections.

Let $X\in\mathbb{R}^{m\times n\times d\times c_{in}}$ be a tensor which represents data with three spatial dimensions, $m\times n\times d$, and $c_{in}$ channels. For example, $X$ could represent a 3D MRI scan $(m=n=d=128)$ with four channels $(c_{in}=4)$.
Then $\mathcal{F}_\theta:\mathbb{R}^{m\times n\times d\times c_{in}}\rightarrow\mathbb{R}^{m\times n\times d\times c_{out}}$ represents a CNN that takes a 3D tensor, $X$, and produces a voxel-wise segmentation with $c_{out}$ possible classes.
We define the operator, $\vect()$, that `flattens' a tensor into a vector such that for $x = \vect(X)\in\mathbb{R}^{mndc_{in}}$, we have $x_{j+ni+mnk+mndc} = X_{i,j,k,c}$. We likewise define the operator, $\tensor()$, which reverses this operation, ie $\tensor(x) = X$. It is possible to write down an input dependent matrix $A[x]\in\mathbb{R}^{mndc_{out} \times mndc_{in}}$ such that,
\begin{equation}\label{eqn:cnn_as_matrix}
    A[x] x = \vect(\mathcal{F}_\theta(X)).
\end{equation}
We direct readers to \cite{baranuik} for a precise construction.
We will refer to the left hand side as the `matrix representation' which acts over vectors. We will refer to the right hand side as the `tensor representation' which acts over tensors. In order to have a fixed matrix to analyze, we compute the SVD of $A[x]$ \textit{for each image} $x$.

While it is theoretically possible to explicitly create the matrix, it may not be practical. Therefore, we focus on Krylov subspace methods to compute the SVD since these methods only require operators that compute the products $A[x] y$ and $A^T[x] z$ for any $y\in\mathbb{R}^{mndc_{in}}$ and $z\in\mathbb{R}^{mndc_{out}}$.
Our primary theoretical contribution is in proving that the matrix, $A^T[x]$, which we refer to as the adjoint matrix, corresponds to a closely related CNN, which we refer to as the Adjoint CNN, where,
\begin{equation}\label{eqn:adj_cnn_as_matrix}
\begin{gathered}
    A^T[x] x = \vect(\mathcal{G}_\theta(X)), \\
    \mathcal{G}_\theta:\mathbb{R}^{m\times n\times d\times c_{out}}\rightarrow\mathbb{R}^{m\times n\times d\times c_{in}}.
\end{gathered}
\end{equation}
We outline our theoretical construction of the Adjoint CNN in Section \ref{methods:theory}. Following this in Section \ref{methods:implementation_solver}, we discuss how we implement SVD solvers and how the computed singular triplets can be used to create a `low rank' approximation of the CNN. Then in Section \ref{methods:mnist} we present the methods using the MNIST dataset to verify properties of the SVD. Next, in Section \ref{methods:brats} we present the methods behind our analysis of CNNs applied to tumor segmentation with the BraTS dataset. 

\subsection{Theory to Construct the Adjoint CNN}\label{methods:theory}

Each operation in a CNN can be written down as a (possibly input-dependent) matrix. This enables us to write down,
\begin{equation}\label{eqn:tensor_vs_matrix_rep_as_product}
    \mathcal{F}_\theta(X) = F_{L} \circ F_{L-1}\circ \dots \circ F_0 (X) \leftrightarrow A[x] = A_L[x]A_{L-1}[x]\dots A_0[x],
\end{equation}
where each $F_\ell$ corresponds to one operation.
Therefore, we can compute the adjoint matrix, and consequently the adjoint CNN, by identifying what $A_\ell^T[\cdot]$ performs for each operation. To provide an example, we will analyze downsampling and upsampling. Then we will state the results for the other operations in Table \ref{tab:summary_of_operations_adjoint} and direct the reader to the supplemental information for their formal proofs.

The CNNs in this paper use sumpooling for downsampling and nearest neighbor for upsampling. This choice was made for two reasons: the associated matrices are input independent and are adjoint to each other. However, is possible to use maxpooling instead at the cost of introducing more input-dependent components. Before defining sumpooling and nearest neighbor upsampling, we define the region associated with a voxel.
\begin{definition}[Region]
    Given a voxel located at $(i,j,k)\in [m]\times [n]\times [d]$,\footnote{$[m] = \{0,1,\dots,m-1\}$} its (tensor) region is:
        \begin{equation*}
            \begin{split}
                R_{(i,j,k)} &= \{ (2i+\delta_0, 2j+\delta_1, 2k+\delta_2 | \delta_0,\delta_1,\delta_2 \in \{0,1\}\}, \\
                    &\subset [2m] \times [2n] \times [2d].
            \end{split}
        \end{equation*}

        For the corresponding flattened index, $p=j+ni+mnk$, we overload the $\vect(\cdot)$ function to define its (vector) region as:
        \begin{equation*}
        \begin{split}
            R_{(p)} &= \text{vec}(R_{(i,j,k)}), \\
                &= \{ \widehat{j}+n\widehat{i}+mn\widehat{k}| (\widehat{i},\widehat{j},\widehat{k})\in R_{(i,j,k)}   \}.
        \end{split}
        \end{equation*}
\end{definition}

Assume that $m$, $n$, and $d$ are all even. Then, sumpooling is defined as:
\begin{equation*} 
\begin{gathered}
    \Pi_\downarrow:\mathbb{R}^{m\times n\times d}\rightarrow\mathbb{R}^{\frac{m}{2}\times \frac{n}{2} \times \frac{d}{2}}, \\
    [\Pi_\downarrow(X)]_{i,j,k} = \sum_{(s,t,u)\in R_{(i,j,k)}} X_{s,t,u}.
\end{gathered}
\end{equation*}

\begin{lemma}\label{lem:sumpool}
    The sumpooling operator $\Pi_\downarrow: \mathbb{R}^{m\times n\times d}\rightarrow\mathbb{R}^{\frac{m}{2}\times \frac{n}{2} \times \frac{d}{2}}$ can be expressed as an input independent matrix, $A_\downarrow \in\mathbb{R}^{\frac{mnd}{8}\times mnd}$. Let $e_q\in\mathbb{R}^{mnd}$ represent the $q$th canonical basis vector. Then the matrix has rows defined as,
    \begin{equation*}
        \begin{gathered}
            \relax [A_{\downarrow}]_{p,\cdot} = \sum_{q\in R_{(p)}} e_q^T,
        \end{gathered}
        \end{equation*}
        where,
        \begin{equation*}
            A_{\downarrow} x = \text{vec}\big( \Pi_\downarrow(X)\big).
        \end{equation*}
\end{lemma}

Nearest neighbor upsampling is a closely related operation defined as:
\begin{equation*}
\begin{gathered}
    \Pi_\uparrow:\mathbb{R}^{m\times n\times d} \rightarrow \mathbb{R}^{2m\times 2n\times 2d} \\
    [\Pi_\uparrow(X)]_{s,t,u} = X_{i,j,k},\quad (s,t,u)\in R_{(i,j,k)}.
\end{gathered}
\end{equation*}

\begin{lemma}\label{lem:nn_up}
    The nearest neighbor upsampling operator $\Pi_\uparrow: \mathbb{R}^{m\times n\times d}\rightarrow\mathbb{R}^{2m\times 2n \times 2d}$ can be expressed as an input independent matrix, $A_\uparrow \in\mathbb{R}^{8mnd\times mnd}$, with columns defined as,
    \begin{equation*}
        \begin{gathered}
            \relax [A_{\uparrow}]_{\cdot,p} = \sum_{q\in R_{(p)}} e_q,
        \end{gathered}
        \end{equation*}
        such that,
        \begin{equation*}
            A_{\uparrow} x = \text{vec}\big( \Pi_\uparrow(X)\big).
        \end{equation*}
\end{lemma}

\begin{lemma}\label{lem:adj_sum_nn}
    Assume $m,n,d$ even. Sumpooling, $\Pi_\downarrow:\mathbb{R}^{m\times n\times d}\rightarrow\mathbb{R}^{\frac{m}{2}\times \frac{n}{2} \times \frac{d}{2}}$, and nearest neighbor upsampling, $\Pi_\uparrow:\mathbb{R}^{\frac{m}{2}\times \frac{n}{2} \times \frac{d}{2}}\rightarrow\mathbb{R}^{m\times n\times d}$, are adjoint to each other.
\end{lemma}
\textit{Proof - }Use Lemma \ref{lem:sumpool}, and consider the $p$th column of $A_\downarrow^T$:
\begin{equation*}
\begin{split}
    [A_\downarrow^T]_{\cdot,p} &= \big[[A_\downarrow]_{p,\cdot}\big]^T \\
        &= \bigg( \sum_{q\in R_{(p)}} e_q^T\bigg)^T = [A_\uparrow]_{\cdot,p}
\end{split}
\end{equation*}

%%% Table with operations and their adjoint
\begin{table}[ht]
    \centering
    \begin{tabular}{cccc}
        Operation & Example & Adjoint & Example \\\hline
        Sumpooling & $\begin{bmatrix}
            1 & -5 \\ 3 & 2
        \end{bmatrix} \mapsto [1]$ & Nearest Neighbor Upsampling & $[1] \mapsto \begin{bmatrix}
            1 & 1 \\ 1 & 1
        \end{bmatrix}$ \\
        Maxpooling & $\begin{bmatrix}
            1 & -5 \\ \textbf{3} & 2
        \end{bmatrix} \mapsto [3]$ & Zero stuffing & $[3] \mapsto \begin{bmatrix}
            0 & 0 \\ \textbf{3} & 0
        \end{bmatrix}$ \\
        Convolution & $\mathcal{C}\in\mathbb{R}^{3\times 3\times 3\times \alpha \times \beta}$ & Convolution & $\widetilde{\mathcal{C}}\in\mathbb{R}^{3\times 3\times 3\times \beta\times\alpha}$ \\
        &&&$\widetilde{\mathcal{C}}_{i,j,k,t,s} = \mathcal{C}_{-i,-j,-k,s,t}$ \\
        ReLU & $\begin{bmatrix}
            1 & -5 \\ 3 & 2
        \end{bmatrix} \mapsto \begin{bmatrix}
            1 & 0 \\ 3 & 2
        \end{bmatrix}$ & `Frozen' ReLU & $\begin{bmatrix}
            a & b \\ c & d
        \end{bmatrix} \mapsto \begin{bmatrix}
            a & 0 \\ c & d
        \end{bmatrix}$ \\
        Skip Connection & $x\mapsto \begin{bmatrix}
            Mx \\ x
        \end{bmatrix}$ & Residual Connection & $\begin{bmatrix}
            y_\alpha \\ y_\beta 
        \end{bmatrix} \mapsto My_\alpha + y_\beta$
    \end{tabular}
    \caption{Summary of CNN operations and the operation associated with their adjoint.}
    \label{tab:summary_of_operations_adjoint}
\end{table}

Using Table \ref{tab:summary_of_operations_adjoint} and Equation \ref{eqn:tensor_vs_matrix_rep_as_product}, we may construct the Adjoint CNN:
\begin{equation}\begin{gathered}
    A^T[x] = A_0^T[x]\dots A_{L-1}^T[x] A_L^T[x] %= B_L[x] B_{L-1}[x]\dots B_0[x]
    \\ \leftrightarrow \\
    \mathcal{G}_\theta(X) = G_{L} \circ G_{L-1}\circ \dots \circ G_0 (X).
\end{gathered}
\end{equation}

\subsection{Implementation of Solver and its Utility}\label{methods:implementation_solver}

We first present how we use the SVD to create a `low rank CNN'. Assume $A[x]$ has rank $r$ and SVD:
%Let $\mathcal{F}_\theta:\mathbb{R}^{m\times n\times d\times c_{in}}\rightarrow \mathbb{R}^{m\times n\times d\times c_{out}}$ represent the CNN and let $A[x]$ be the associated rank $r$ matrix with SVD:

\begin{equation}\label{eqn:svd_as_sum}
    A[x] = \sum_{i=1}^r \sigma_i[x] u_i[x] v_i^T[x].
\end{equation}
Traditionally, the CNN generates class probabilities by computing:
\begin{equation}
\begin{split}
    \widehat{P}(X) &= \sigma_{\text{softmax}}\big(\mathcal{F}_\theta(X)\big)\in \mathbb{R}^{m\times n\times d\times c_{out}},
\end{split}
\end{equation}
where the final predicted segmentation is:
\begin{equation}\label{eqn:label_from_argmax}
\begin{gathered}
    P(X)\in\mathbb{R}^{m\times n\times d}, \\
    [P(X)]_{i,j,k} = \argmax_{c}[\widehat{P}(X)]_{i,j,k,c} \in\{0,1,\dots,c_{out}-1\}.
\end{gathered}
\end{equation}

Notice that the CNN model class probabilities could be equivalently computed as:
\begin{align*}
    \widehat{P}(X) &= \sigma_{\text{softmax}}\Big(\mathcal{F}_\theta(X)\Big) \\
        &= \sigma_{\text{softmax}}\Big(\tensor(A[x] x)\Big) \\
        &= \sigma_{\text{softmax}}\Big(\tensor(\sum_{i=1}^r \sigma_i[x] u_i[x] v_i^T[x] x)\Big).
\end{align*}

This insight allows us to compute the prediction from the `$k$ rank CNN' by using the $k$ rank approximation of $A[x]$, denoted $A_k[x]$, as:
\begin{equation}\label{eqn:k_rank_CNN_prediction}
\begin{split}
    \widehat{P_k}(X) &= \sigma_{\text{softmax}}\Big(\tensor(\sum_{i=1}^k \sigma_i[x] u_i[x] v_i^T[x] x)\Big), \\
        &= \sigma_{\text{softmax}}\Big(\tensor( A_k[x] x )\Big).
\end{split}
\end{equation}

%We reiterate that we compute the SVD of $A[x]$ for a fixed $x$. This means for each image $x$, the SVD solver will require the matrix vector products $A[x] z_0$ for any $z_0 \in\mathbb{R}^{n}$ and $A^T[x] z_1$ for any $z_1\in\mathbb{R}^m$. Since the only input dependent operation we used for the models in this paper was ReLU, this is implemented by replacing each ReLU from both the CNN and Adjoint CNN with a binary tensor that captures how the ReLU was activated for the input $x$. Details how how this was accomplished can be found in Appendix \ref{app:freeze_relu}.

We use the Python adaptation of SLEPc \cite{slepc4py1,slepc4py2} to compute the SVD. We use its default ``cross'' solver. This solver computes the eigenvalue decomposition of $H(A) = A^T A = V \Lambda V^{-1}$ using an implicit Krylov-Schur method to compute the singular values and right singular vectors. Then it computes the left singular vectors by normalizing $Av_j$.
SLEPc evaluates the error of its approximation by computing the residual,
\begin{equation*}
    ||r||_2 = \big( ||A\widetilde{v} - \widetilde{\sigma} \widetilde{u}||_2^2 + ||A^T\widetilde{u} - \widetilde{\sigma}\widetilde{v}||_2^2 \big)^{\frac{1}{2}},
\end{equation*}
where $(\widetilde{\sigma},\widetilde{u},\widetilde{v})$ represents the solver's estimate for an arbitrary singular triplet. We set SLEPc to consider the relative error.

In addition to computing the rank $k$ CNN prediction using Equation \eqref{eqn:k_rank_CNN_prediction}, we are interested in the subspaces formed by the span of the computed singular vectors.
Recall the singular vectors are orthonormal and hence we can consider the projection of a sample, $x$, onto the span of the right singular vectors:
\begin{equation}
    V_k[x] V_k^T[x] x, \quad V_k[x] = [v_1[x] \dots v_k[x]].
\end{equation}
We use the following ratio, which we call the Right Projection Ratio (RPR), to evaluate how well $x$ is approximated by its projection onto the subspace spanned by the $k$ right singular vectors:
\begin{equation}\label{eqn:RPR}
    \text{RPR}(x;k) = \frac{||V_k[x] V_k^T[x] x||_2^2}{||x||_2^2} \in[0,1],
\end{equation}
where,
\begin{equation*}
    % \frac{||V_k[x] V_k^T[x] x||_2^2}{||x||_2^2}
    \text{RPR}(x;k) = \begin{cases}
        1, & x\in R(A_k^T[x]) \\
        0, & x\in N(A_k[x])\supset N(A[x])
    \end{cases}
\end{equation*}
We may use this ratio as an estimate for how close the input image is to the nullspace of the CNN. To illustrate why an RPR close to zero is problematic, assume $x\in N(A[x])$. Then,
\begin{equation}
    x\in N(A[x]) \implies \mathcal{F}_\theta(X) \equiv 0\in\mathbb{R}^{m\times n \times d\times c_{out}}.
\end{equation}
To focus on an arbitrary voxel, $(i,j,k)$, notice that the class probabilities are,
\begin{equation}
\begin{split}
    [\widehat{P}(X)]_{i,j,k,\cdot} &= [\sigma_{\text{softmax}}(0)]_{i,j,k,\cdot} \\
        &= [\frac{1}{c_{out}},\dots,\frac{1}{c_{out}}].
\end{split}
\end{equation}
This means that each class is equally likely to be selected and represents a high degree of model uncertainty. Additionally, this could introduce algorithmic bias. For example, the argmax functions in both NumPy and PyTorch always return the indices of the first occurrence when there are multiple occurrences of the maximum value.

It is a little more involved to evaluate a similar ratio using the left singular vectors. For this discussion, we consider a CNN tasked with image classification. 
We need to construct a mapping from the space of possible labels to the raw logits output space.
Assume $x$ has label $\ell\in\{0,1,\dots,c_{out}-1\}$. From Equation \eqref{eqn:label_from_argmax}, we see that we need to pick $y\in\mathbb{R}^{c_{out}}$ such that $\ell=\argmax(y)$. Once a mapping is created, we can compute the Left Projection Ratio (LPR) which evaluates how well the projection of $y$ onto the $k$ left singular vectors approximates $y$:

\begin{equation}\label{eqn:LPR}
    \text{LPR}(y;k) = \frac{||U_k[x] U_k^T[x] y||_2^2}{||y||_2^2} \in[0,1].
\end{equation}
Similar to the RPR, we have,
\begin{equation*}
% \frac{||U_k[x] U_k^T[x] y||_2^2}{||y||_2^2}
    \text{LPR}(y;k) = \begin{cases}
        1, & y\in R(A_k[x])\subset R(A[x]) \\
        0, & y\in N(A_k^T[x])\supset N(A^T[x])
    \end{cases}
\end{equation*}

\subsection{Image Classification}\label{methods:mnist}

The MNIST dataset consists of 60000 training images and 10000 testing images. Each image, $X\in\mathbb{R}^{28\times 28}$, belongs to exactly one class out of 10 possible. Therefore, the matrix representation of this problem has,
\begin{equation*}
    A[x]\in\mathbb{R}^{10\times 28^2}.
\end{equation*}
With an upper bound known for the rank, we compute the rank $k\in\{1,2,\dots,10\}$ approximation for each image.

The CNN architecture consists of the following layers:
\begin{itemize}
    \item Convolution with kernel size $5\times 5$, 32 filters, zero padding, followed by ReLU,
    \item Sumpooling,
    \item Convolution with kernel size $5\times 5$, 64 filters, zero padding, followed by ReLU,
    \item Sumpooling,
    \item A linear mapping to 512 nodes followed by ReLU,
    \item A linear mapping to 256 nodes followed by ReLU,
    \item A linear mapping to 10 nodes,
    \item Softmax final activation.
\end{itemize}
We use a batch size of 32, stochastic gradient descent with cross entropy loss, a learning rate of $3\times 10^{-3}$, and we train for 50 epochs.

We use the default SVD solver from SLEPc4py and set the tolerance on the relative error of the solver to $10^{-5}$.

As an additional experiment to determine if the SVD can be used to identify bias in the model, we train a second CNN on a heavily unbalanced training set. For the unbalanced training set, we remove $99\%$ of the images belonging to digits $\{1, 9\}$.

In order to evaluate the LPR, we map the label $\ell\in\{0,1,\dots,9\}$ to $y\in\mathbb{R}^{10}$ where,
\begin{equation*}
    [y(\ell)]_i = \begin{cases}
        2, & i=\ell \\
        -2, & i\neq \ell
    \end{cases}
\end{equation*}

This choice was made by observing that after applying softmax, we have:
\begin{equation*}
    \sigma_{\text{softmax}}(y(\ell))_i \approx \begin{cases}
        0.86, & i=\ell \\
        0.016, & i\neq \ell
    \end{cases}
\end{equation*}
Broadly interpreted, this mapping is one way to check if the model can predict with $86\%$ certainty the correct class. This selection was motivated by observing that the rank 1 Matrix Representation assigned an average probability around $0.8$ when it predicted correctly and that the Tensor Representation assigned an average probability nearly equal to $1$ for its correct predictions.

\subsection{Image Segmentation}\label{methods:brats}

To consider a more practical application where it would be infeasible to compute the compact SVD, we demonstrate as a proof of concept how to analyze a CNN trained to segment brain tumors.
We use the Brain and Tumor Segmentation 2020 dataset from \cite{brats1,brats2,brats3} which consists of 369 3D multi-modal MRI scans with multi-label tumor segmentation masks. We combine the tumor masks to consider binary tumor segmentation. We use the MIST framework from \cite{mist_paper,mist_repo} to preprocess the data and perform five fold cross validation with a general UNet based architecture:
\begin{itemize}
    \item Operates on patches of size $(128,128,128)$,
    \begin{itemize}
        \item That is, $\mathcal{F}_\theta:\mathbb{R}^{128\times 128\times 128\times 4}\rightarrow\mathbb{R}^{128\times 128\times 128 \times 2}$
    \end{itemize}
    \item Depth 5,
    \item Each convolutional layer consists of two convolution+ReLU pairs,
    \begin{itemize}
        \item We use the Pocket Paradigm from \cite{mist_paper} and fix the number of convolution filters to 32
    \end{itemize}
    \item Sumpooling,
    \item Nearest neighbor upsampling,
    \item Skip connections
\end{itemize}
We use a batch size of 2, the Adam optimizer with L2 relaxed dice loss, a learning rate of $3\times 10^{-4}$, and train for 1000 epochs. The validation set was used to prevent overfitting and select the best model.

We use the default SVD solver from SLEPc4py and set the tolerance on the relative error to $10^{-4}$.
We chose to compute 10 singular triplets.

For both training and computing the SVD, we used Monai's sliding window inference function to stitch together the full prediction from patches.

\section{Results}\label{sec:results}

In Section \ref{results:mnist} we present the image classification results with the MNIST dataset.
In Section \ref{results:brats}, we present our findings from the segmentation problem with the BraTS dataset.

\subsection{MNIST Classification}\label{results:mnist}

\begin{figure}[t!]
    \centering
    \begin{subfigure}[t]{0.49\textwidth}
        \centering
        \includegraphics[width=1\linewidth]{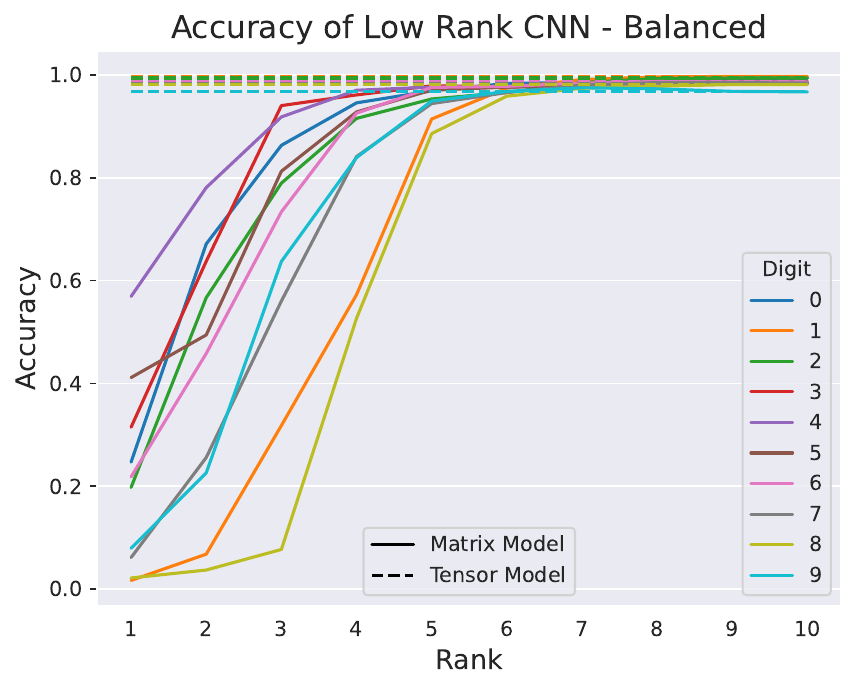}
    \end{subfigure}%
    ~
    \begin{subfigure}[t]{0.49\textwidth}
        \centering
        \includegraphics[width=1\linewidth]{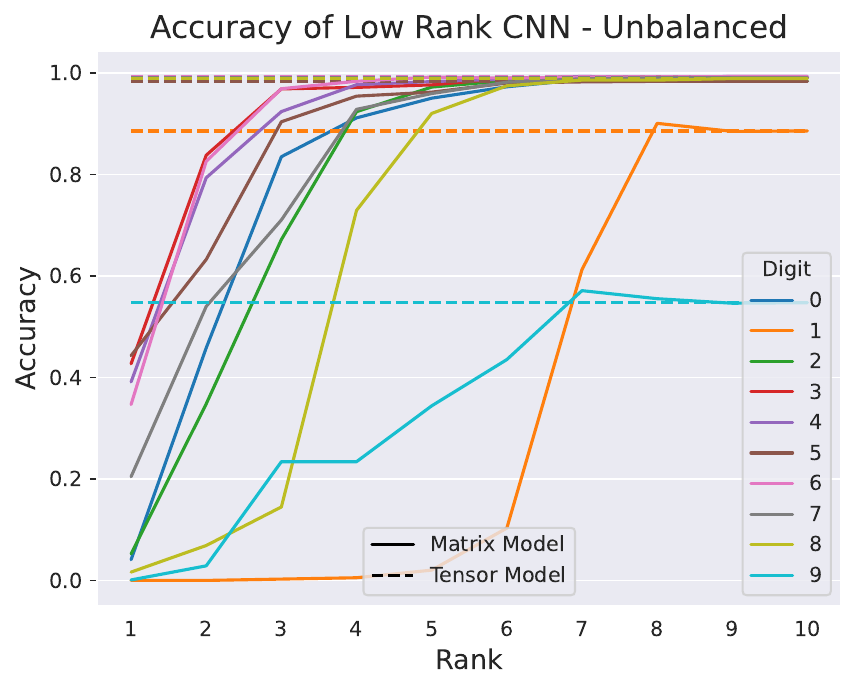}
    \end{subfigure}
    \caption{Evaluating the digit specific accuracy of the balanced (left) and unbalanced (right) rank $k$ CNN.}
    \label{fig:mnist_digit_accuracy}
\end{figure}

We evaluated both CNNs, referred to as the `Tensor Model', on the test set. The well balanced Tensor Model correctly labeled 9864 images giving it an accuracy of $0.9864$ and the unbalanced Tensor Model correctly labeled 9329 images giving it an accuracy $0.9329$.
It is important to highlight that the balanced rank 10 CNN also had an accuracy of $0.9864$ and failed on the same 136 images as the balanced Tensor Model. Likewise, the unbalanced rank 10 CNN failed on the same  671 images as the unbalanced Tensor Model.
In Figure \ref{fig:mnist_digit_accuracy} we compare the rank $k$ CNN accuracy to the Tensor Model stratified over each digit.

\begin{figure}[t!]
    \centering
    \begin{subfigure}[t]{0.49\textwidth}
        \centering
        \includegraphics[width=1\linewidth]{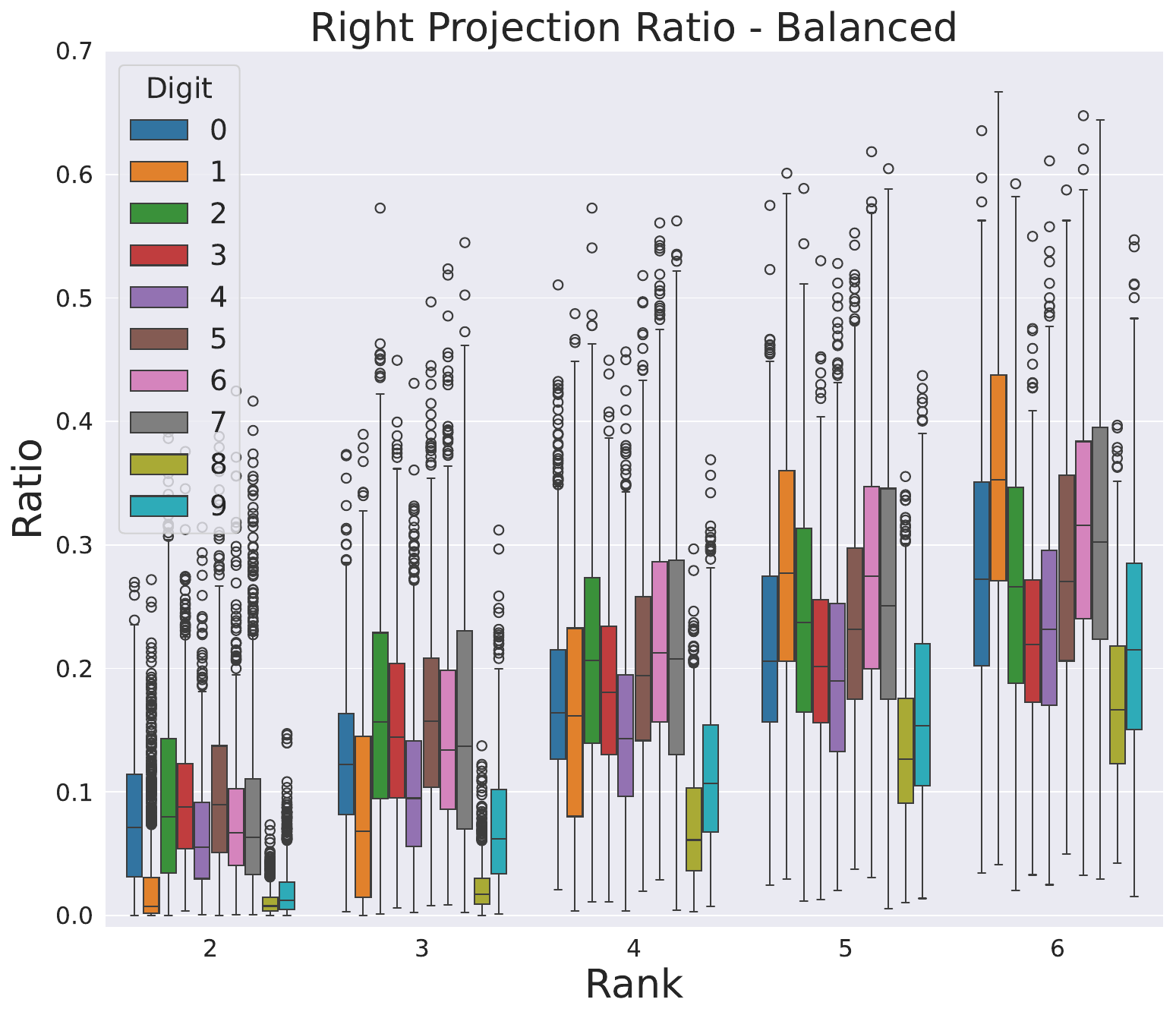}
        \caption{For the CNN trained on the balanced dataset.}
        \label{fig:rpr_balanced}
    \end{subfigure}%
    ~
    \begin{subfigure}[t]{0.49\textwidth}
        \centering
        \includegraphics[width=1\linewidth]{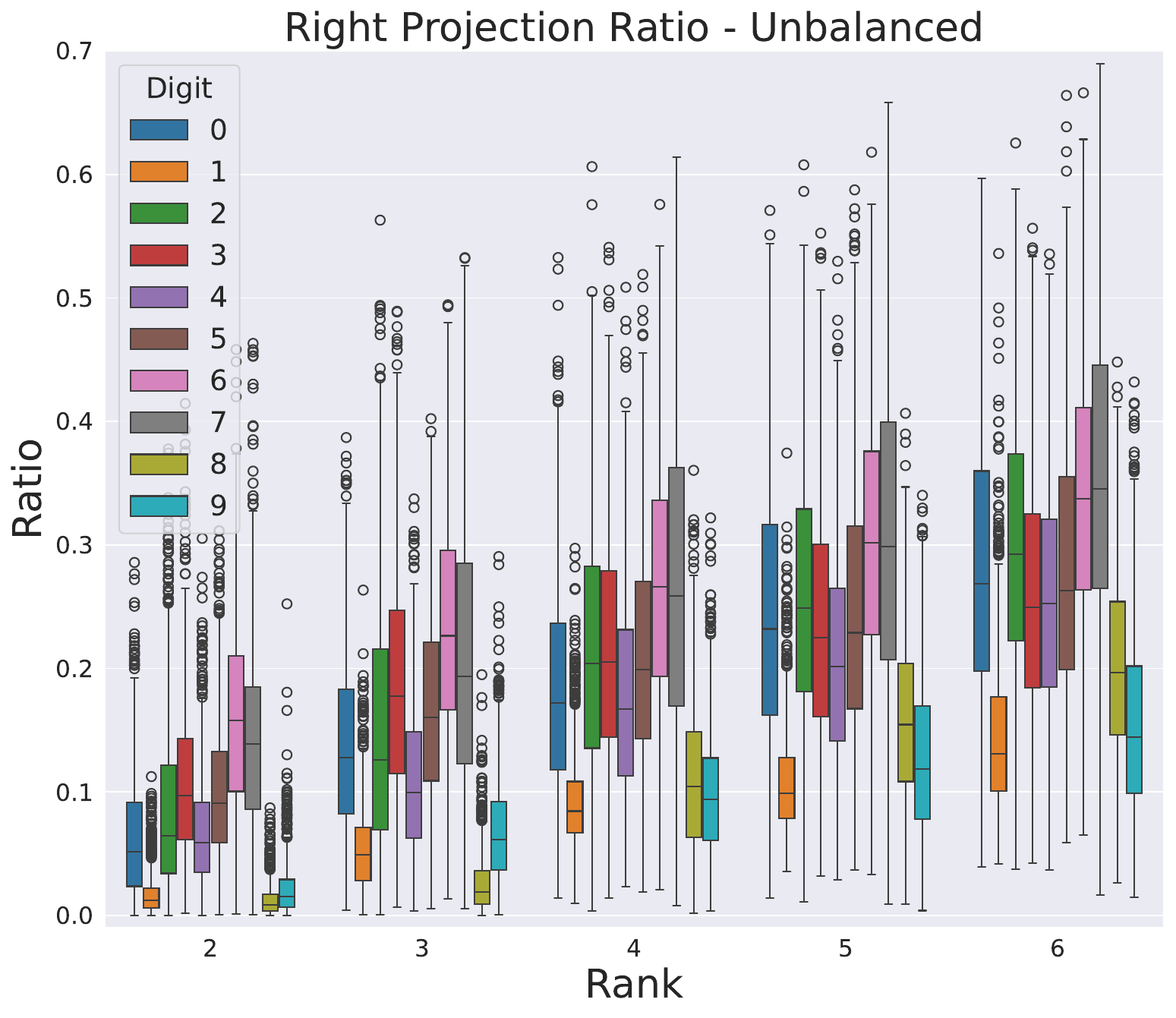}
        \caption{For the CNN trained on the unbalanced dataset.}
        \label{fig:rpr_unbalanced}
    \end{subfigure}
    \caption{Boxplot showing how the RPR changes for each digit as the rank increases. Using the RPRs from the balanced model as a baseline, we observe that digit $1$ deviates the most for the unbalanced dataset. With its significantly stunted RPR, this can be interpreted as images depicting the digit $1$ belonging more to the nullspace of the CNN.
    Additionally, the RPR for digit $9$, which was generally above digit $8$ for the balanced experiment, drops below digit $8$ for the unbalanced experiment. By evaluating proximity to the nullspace with the RPR, we can identify digits $\{1,9\}$ as candidates of poor model performance for the unbalanced experiment.}
    \label{fig:mnist_rpr_rd}
\end{figure}

\begin{figure}[t!]
    \centering
    \begin{subfigure}[t]{0.49\textwidth}
        \centering
        \includegraphics[width=1\linewidth]{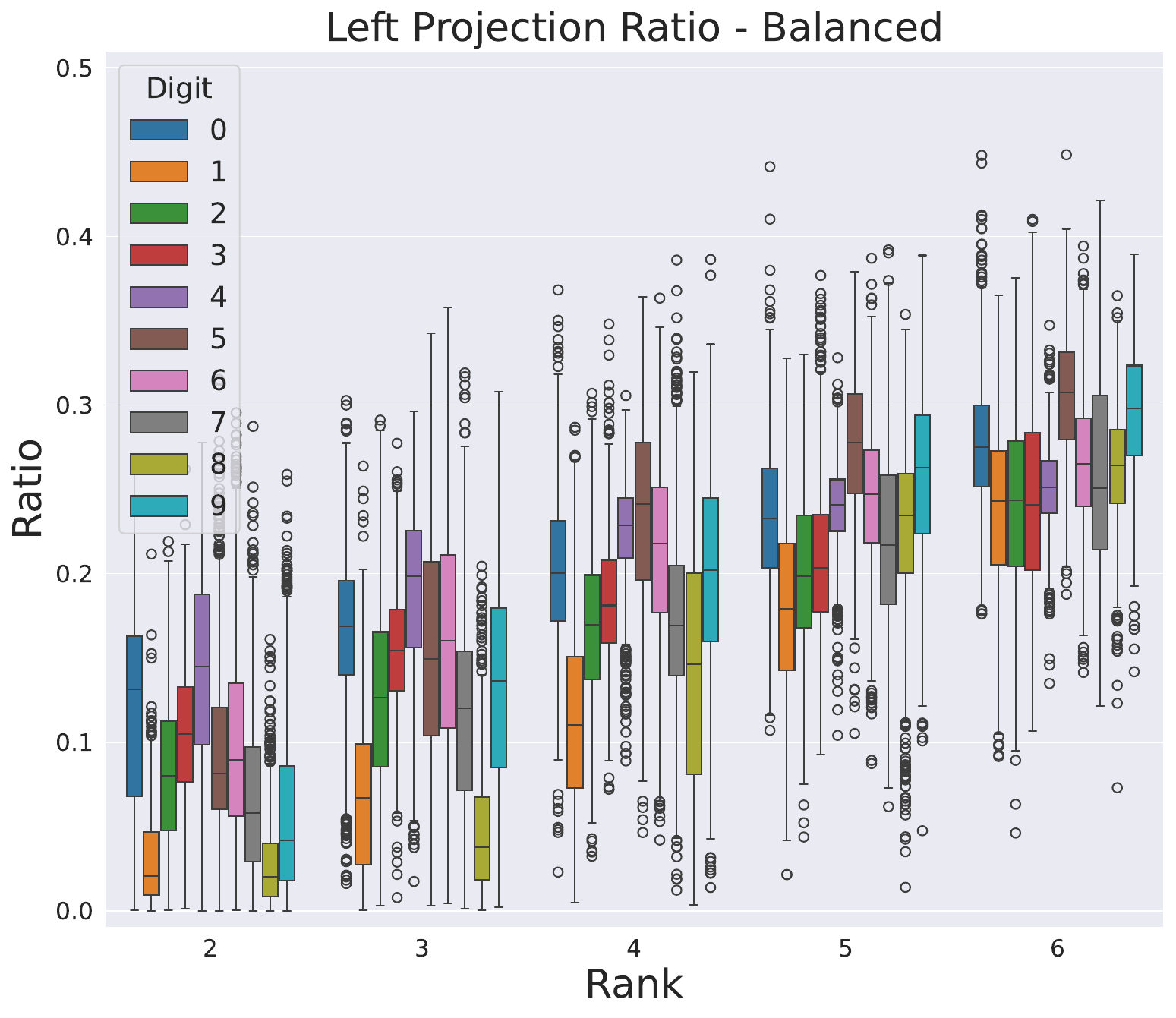}
        \caption{For the CNN trained on the balanced dataset.}
        \label{fig:lpr_balanced}
    \end{subfigure}%
    ~
    \begin{subfigure}[t]{0.49\textwidth}
        \centering
        \includegraphics[width=1\linewidth]{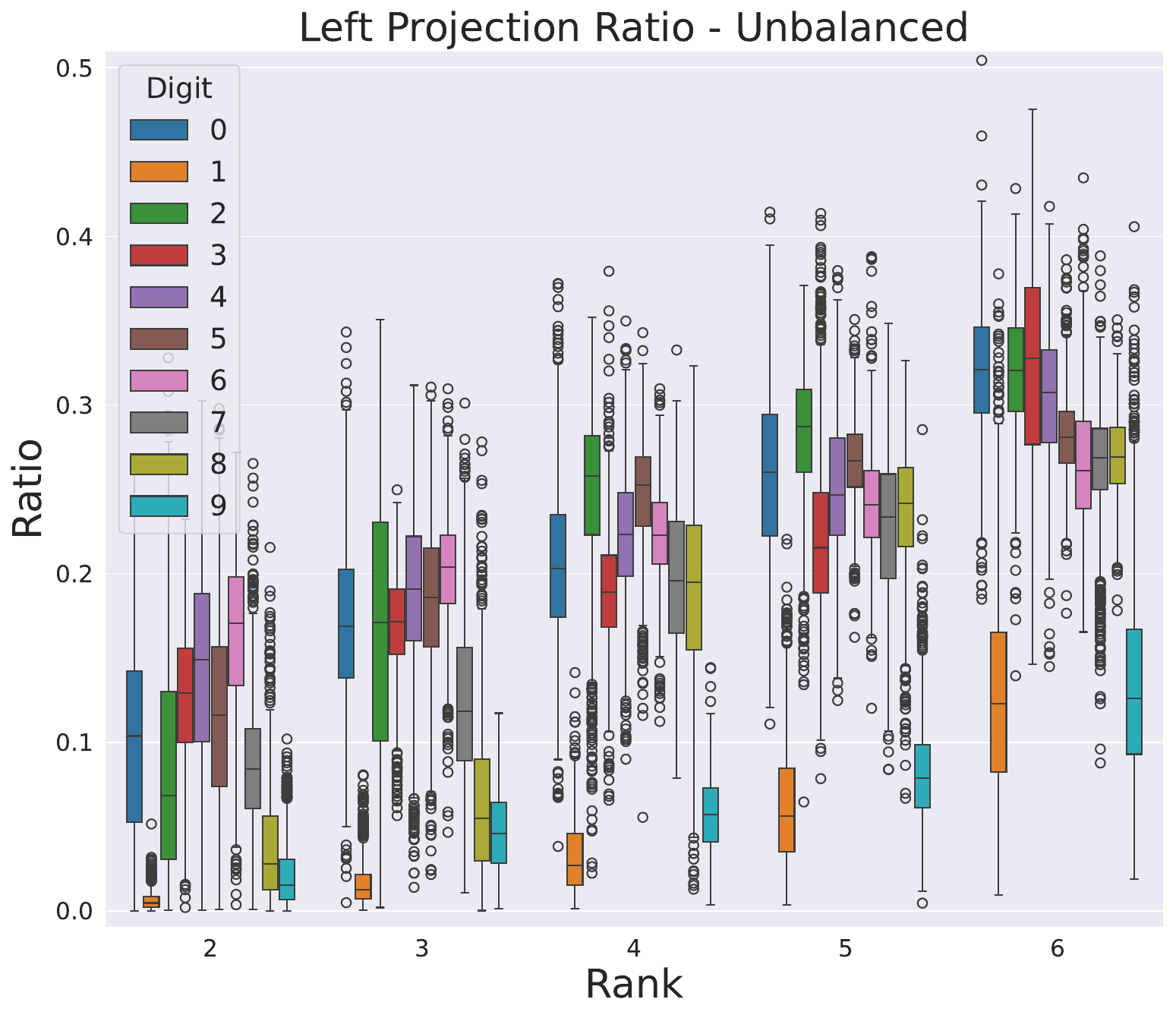}
        \caption{For the CNN trained on the unbalanced dataset.}
        \label{fig:lpr_unbalanced}
    \end{subfigure}
    \caption{Boxplot showing how the LPR changes for each digit as the rank increases. When the training set is unbalanced, digits $\{1,9\}$ consistently have the lowest LPRs. Referring back to Section \ref{methods:implementation_solver}, we recognize that this means these digits are not well represented in the range of the CNN. This offers improved interpretability for why the CNN trained on the unbalanced dataset performed worse than the CNN trained on the balanced dataset.}
    \label{fig:mnist_lpr_rd}
\end{figure}

In order to have class specific information, we look at the distribution of the projection ratios with respect to each digit. We present the distribution at ranks $2, 3, 4, 5, 6$ for the RPR in Figure \ref{fig:mnist_rpr_rd} and for the LPR in Figure \ref{fig:mnist_lpr_rd}.

Finally, included in the Supplemental Information are a few visual examples of the computed singular vectors and the associated projections.

\subsection{BraTS Segmentation}\label{results:brats}

\begin{figure}[ht]
    \centering
    \includegraphics[width=0.75\linewidth]{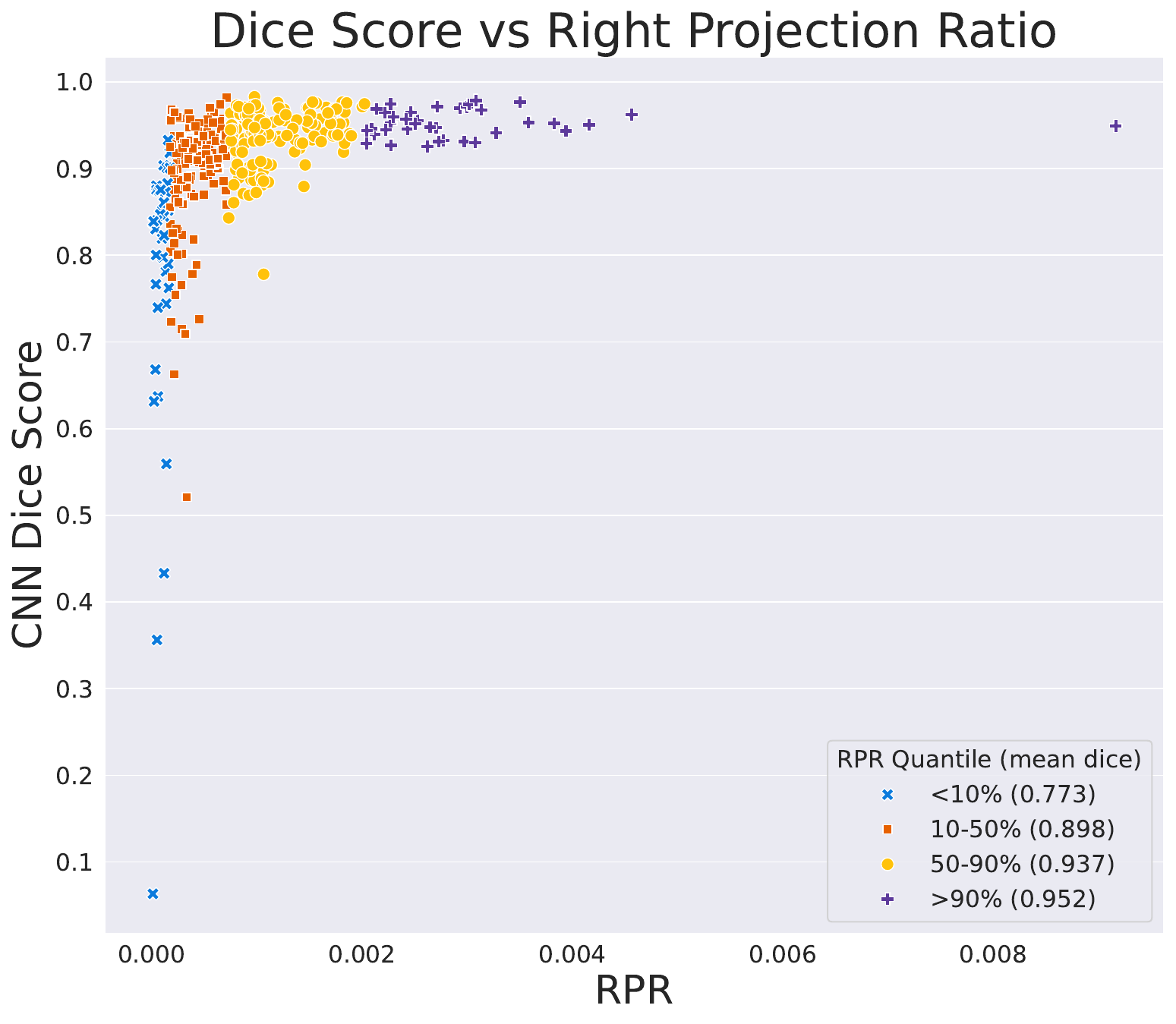}
    \caption{The RPR is calculated using the 10 right singular vectors corresponding to the 10 largest singular values. The `CNN Dice Score' is the dice score from the Tensor Representation of the CNN. Test set contains 369 volumes.}
    \label{fig:brats_dice_vs_rpr_ffx}
\end{figure}

The mean test dice score for the Tensor Representation was $0.9067$. In Figure \ref{fig:brats_dice_vs_rpr_ffx} we present the Tensor CNN dice score against its RPR for every volume. This is summarized in Table \ref{tab:brats_dice_by_rpr_ffx} where we report the mean dice score for images with RPR within certain ranges. 

\begin{table}[ht]
    \centering
    \begin{tabular}{c c}
        RPR Quantile & Mean Dice Score \\\hline
        Below $10\%$ & $0.7728$ \\
        Within $10-50\%$ & $0.8979$ \\
        Within $50-90\%$ & $0.9375$ \\
        Above $90\%$ & $0.9522$ \\\hline
        Entire Set & $0.9067$
    \end{tabular}
    \caption{Breakdown of Mean Dice Score by RPR for full five fold cross validation (369 volumes).}
    \label{tab:brats_dice_by_rpr_ffx}
\end{table}

\section{Discussion}\label{sec:discussion}

We evaluated our SVD-based approach on the MNIST image classification problem since it would be feasible to compute the compact SVD and verify that what we computed as various rank approximations to the CNN does indeed behave like an SVD.
The first validating observation was that for both the CNN trained on the balanced dataset and the CNN trained on the unbalanced dataset, there was steady convergence of the low rank accuracy to the Tensor Model accuracy as shown at the digit level in Figure \ref{fig:mnist_digit_accuracy}. This behavior is consistent with the fact that $A_k[x]$ converges to $A[x]$ as $k$ tends to the full rank $r$. Additionally, the rank 10 matrix model and the Tensor Model failed on the same images which offers empirical support that the computed rank 10 Matrix Representation, $A_{10}[\cdot]$, is fundamentally equivalent to the Tensor Representation, $\mathcal{F}_\theta(\cdot)$.

The second validating observation was that the RPR and LPR monotonically increased with the rank. At rank 10, the LPR was exactly 1. This was expected since the dimension of the label space was 10 and hence $U_{10}[x]$ formed an orthonormal basis that spanned the entire space.

We found that the projection ratios were able to identify the class imbalance between the two CNNs. Looking at the RPR for the balanced CNN in Figure \ref{fig:rpr_balanced} as a baseline, we saw that digit 8 was the slowest digit to improve. Comparing this against the RPR for the unbalanced CNN, we noticed that digits $\{1,8,9\}$ were the slowest to improve. Digit 8 can be seen as an artifact since it was included in both models. This suggests that the RPR provides a means to estimate whether or not there is bias within a model as was the case here.
This signal was even more clear for the LPR in Figure \ref{fig:mnist_lpr_rd} where digits $\{1,9\}$ had significantly smaller ratios.

One might be inclined to explain the poor performance of the unbalanced CNN on digits $\{1,9\}$ in Figure \ref{fig:mnist_digit_accuracy} as a lack of training. With our approach, our explanation is that the CNN was unable to sufficiently adapt its range and null space to properly accommodate the missing digits. That is, the lack of training meant that the digits were not as easy for the model to express. This would require more singular vectors, corresponding to less important features of the CNN, to improve the projection ratios. 

This correlation between the RPR and the CNN performance was additionally observed for tumor segmentation with rank 10 approximations.
When we ordered the test set by the RPR of the volume, we found that the volumes that belonged to the top $10\%$ quantile had an average dice score of $0.9522$ with none below $0.9$. Conversely, the volumes that belonged to the bottom $10\%$ quantile had an average dice score of $0.7728$.
This demonstrates that the RPR could be used as an estimate for the dice score. 
%Additionally, it suggests the RPR could be tailored to identify a coreset.

\section{Conclusion}\label{sec:conclusion}

We introduced a methodology for computing the singular value decomposition of a convolutional neural network. The utility of this computation is in its ability to quantify to what degree an image, a segmentation mask, or a classification label belongs to the range and nullspace of the CNN.
This was accomplished by defining the Right and Left Projection Ratio that use the right and left singular vectors from the truncated SVD respectively.
We observed that the projection ratios were able to identify class imbalance between two models trained for classification with the MNIST dataset.

Additionally, we observed that the Right Projection Ratio can estimate the quality of a segmentation mask for a more challenging task like tumor segmentation. This could be useful in active learning problems where it is desirable to select a batch of images from an unlabeled pool that the CNN would likely perform poorly on. Alternatively, the RPR could be useful in a recommender system where the goal is to select an image that maximizes some objective where the true label or segmentation is unknown.
%Finally, it may be possible to adapt the projection ratios to find a coreset

\subsubsection*{Acknowledgments}

This research was partially funded by a training fellowship from the Keck Center of the Gulf Coast Consortia, on the Training Program in Biomedical Informatics, National Library of Medicine (NLM), PI Lydia Kavraki, grant number: T15LM007093-30.

This work was partly supported by the Tumor Measurement Initiative through the MD Anderson Strategic Initiative Development Program (STRIDE) and QIAC Partnership in Research (QPR) Program. NIH support under R01CA195524, and NSF support under Awards NSF-2111147, 2231482, 2111459 is gratefully acknowledged.
%2231482
%2111459

\bibliography{biblio.bib}
\newpage

\section*{Sample MNIST Projections}

\begin{figure}[ht]
    \centering
    \includegraphics[width=1.\linewidth]{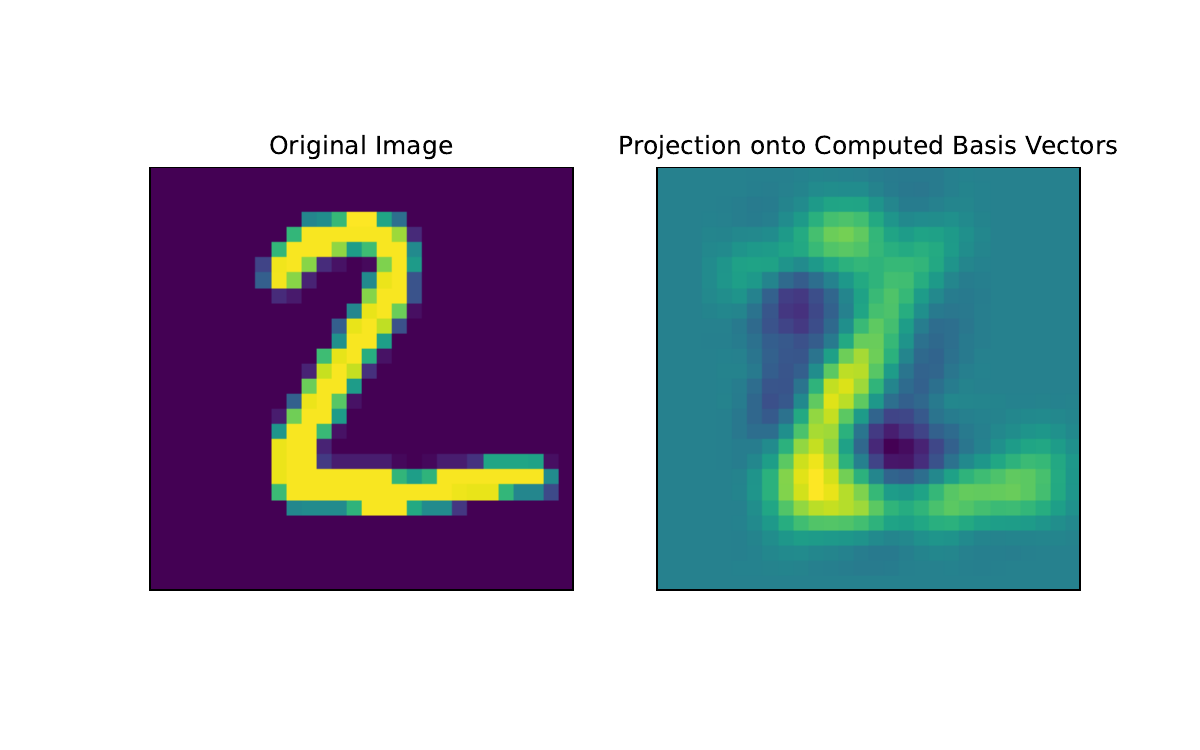}
    \caption{Left: the input image. Right: The projection of the image onto the first 10 singular vectors.}
    \label{fig:projection_sam_1}
\end{figure}
\begin{figure}[ht]
    \centering
    \includegraphics[width=0.48\linewidth]{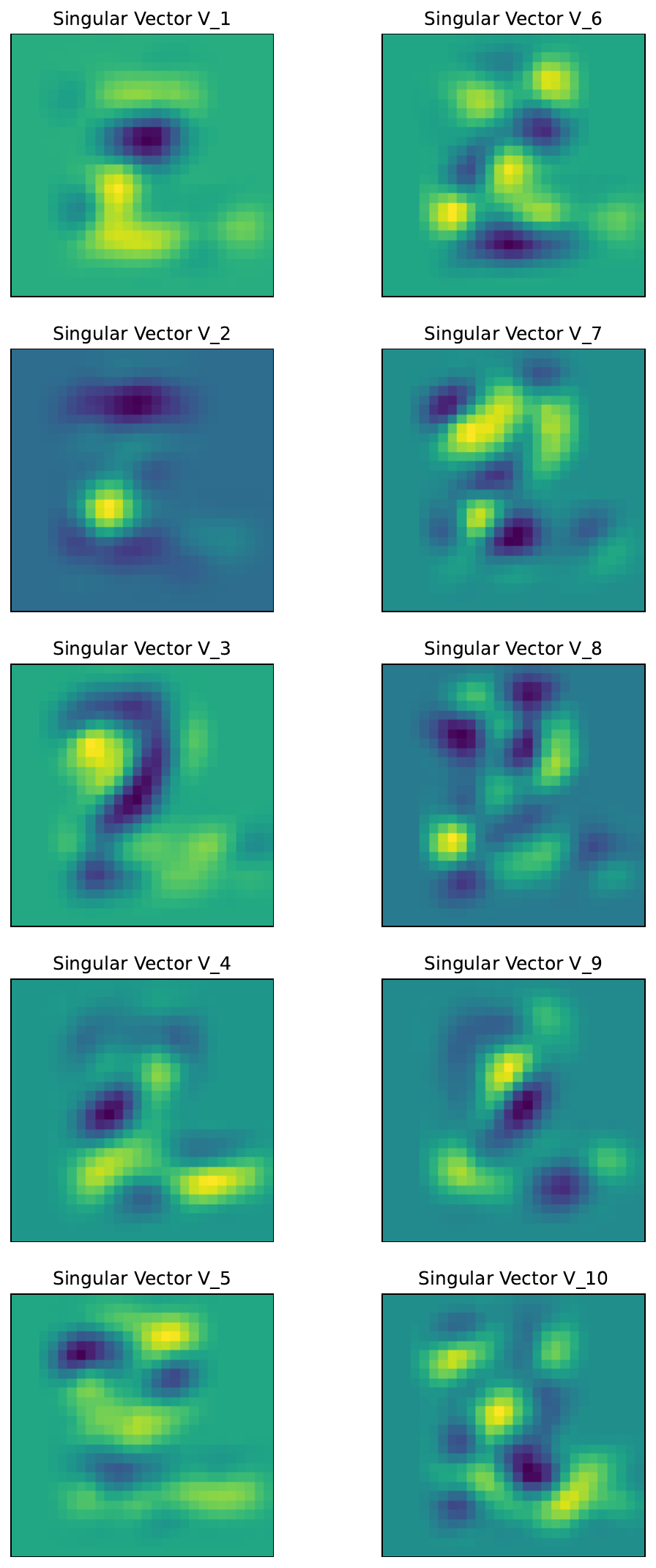}
    \caption{The right singular vectors with respect to the input image in Figure \ref{fig:projection_sam_1}.}
    \label{fig:right_vec_sam_1}
\end{figure}

\begin{figure}[ht]
    \centering
    \includegraphics[width=1.\linewidth]{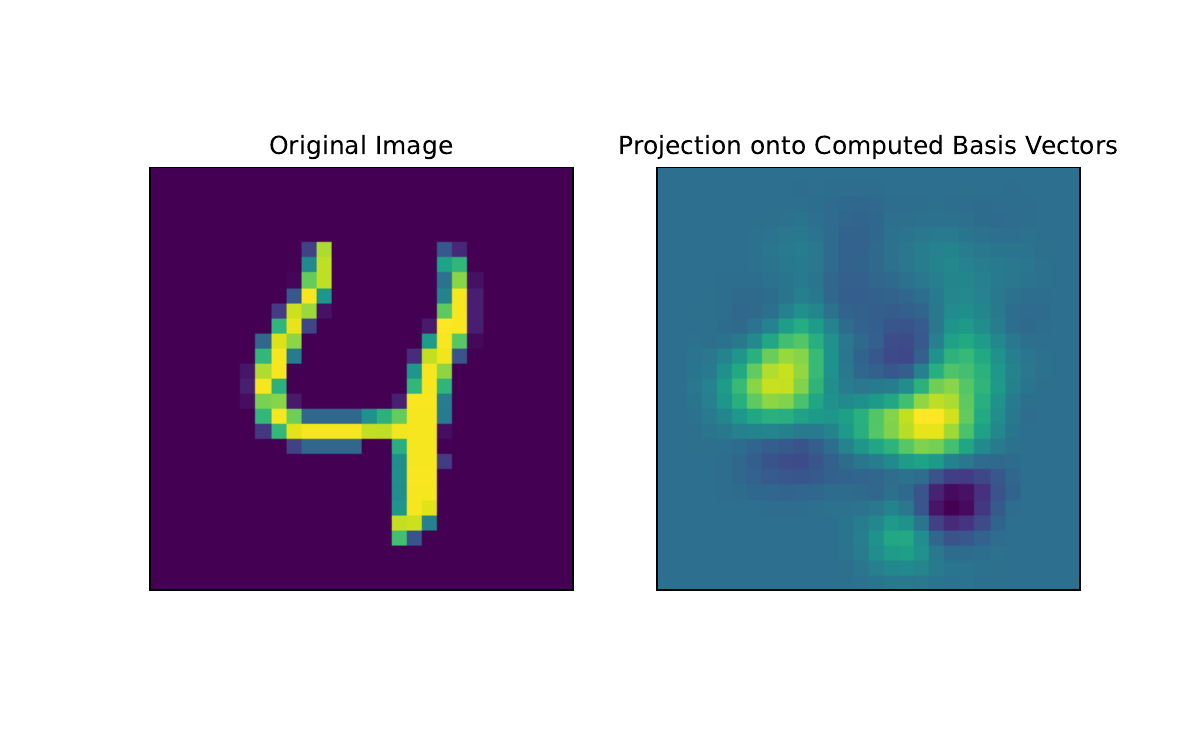}
    \caption{Left: the input image. Right: The projection of the image onto the first 10 singular vectors.}
    \label{fig:projection_sam_4}
\end{figure}
\begin{figure}[ht]
    \centering
    \includegraphics[width=0.48\linewidth]{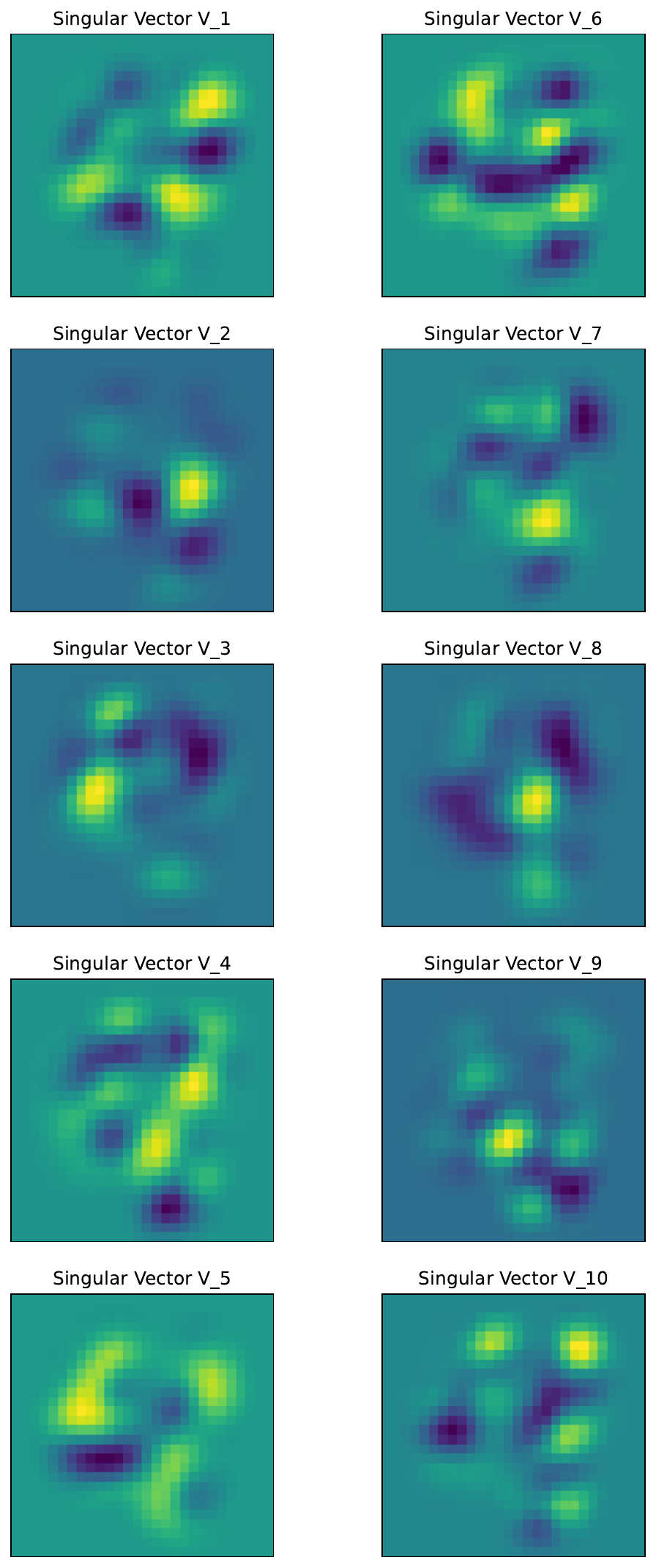}
    \caption{The right singular vectors with respect to the input image in Figure \ref{fig:projection_sam_4}.}
    \label{fig:right_vec_sam_4}
\end{figure}

\begin{figure}[ht]
    \centering
    \includegraphics[width=1.\linewidth]{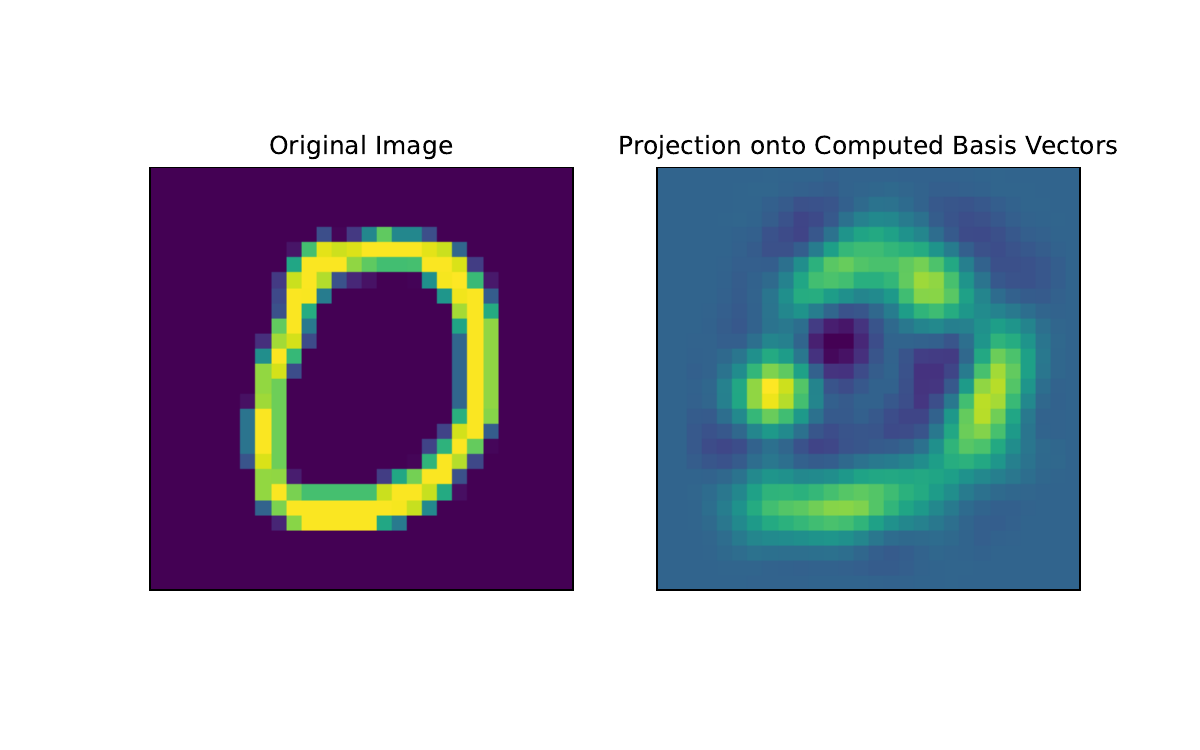}
    \caption{Left: the input image. Right: The projection of the image onto the first 10 singular vectors.}
    \label{fig:projection_sam_10}
\end{figure}
\begin{figure}[ht]
    \centering
    \includegraphics[width=0.48\linewidth]{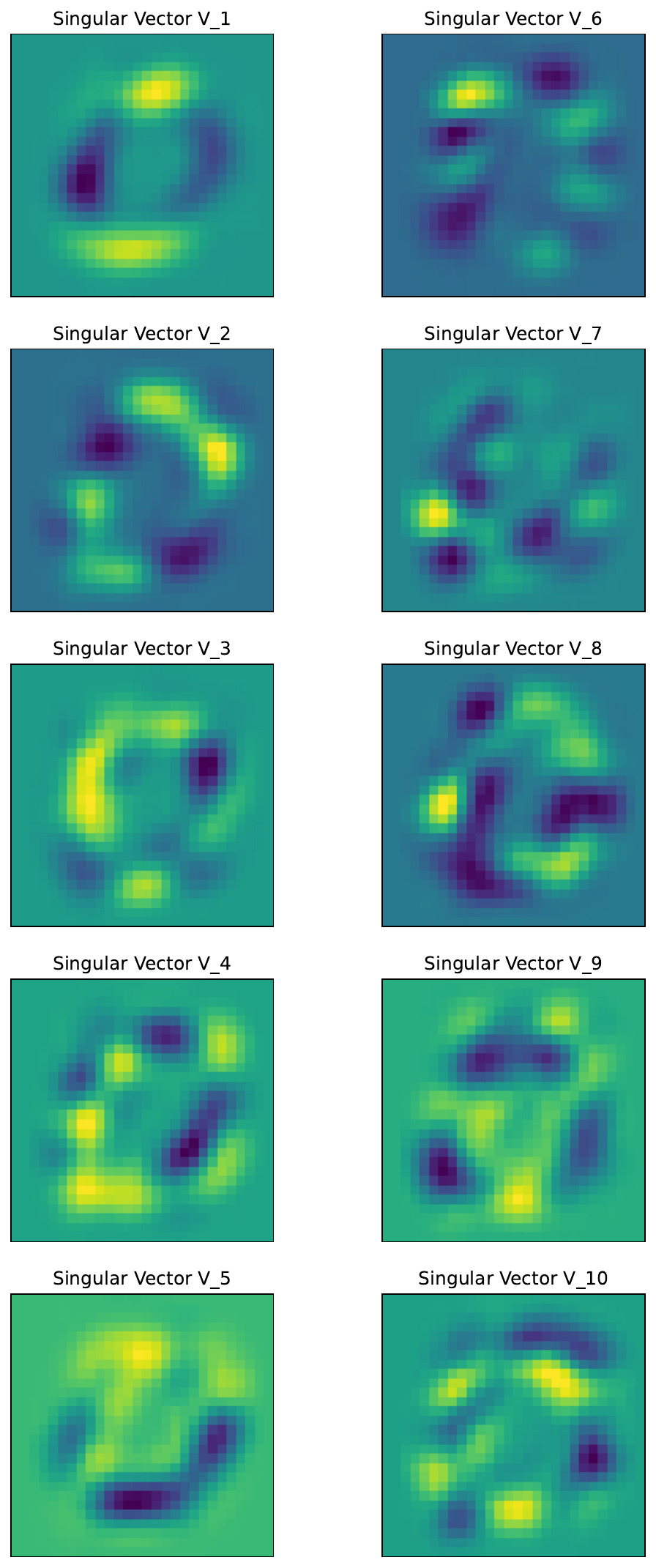}
    \caption{The right singular vectors with respect to the input image in Figure \ref{fig:projection_sam_10}.}
    \label{fig:right_vec_sam_10}
\end{figure}

\begin{figure}[ht]
    \centering
    \includegraphics[width=1.\linewidth]{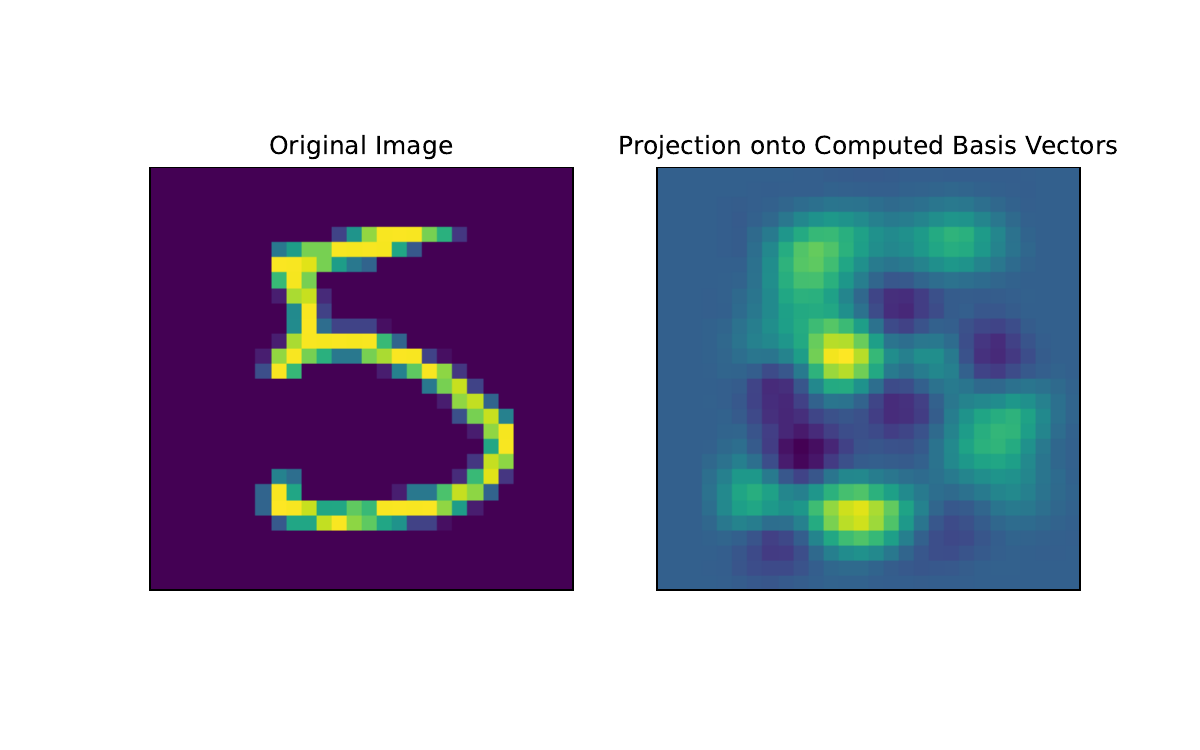}
    \caption{Left: the input image. Right: The projection of the image onto the first 10 singular vectors.}
    \label{fig:projection_sam_15}
\end{figure}
\begin{figure}[ht]
    \centering
    \includegraphics[width=0.48\linewidth]{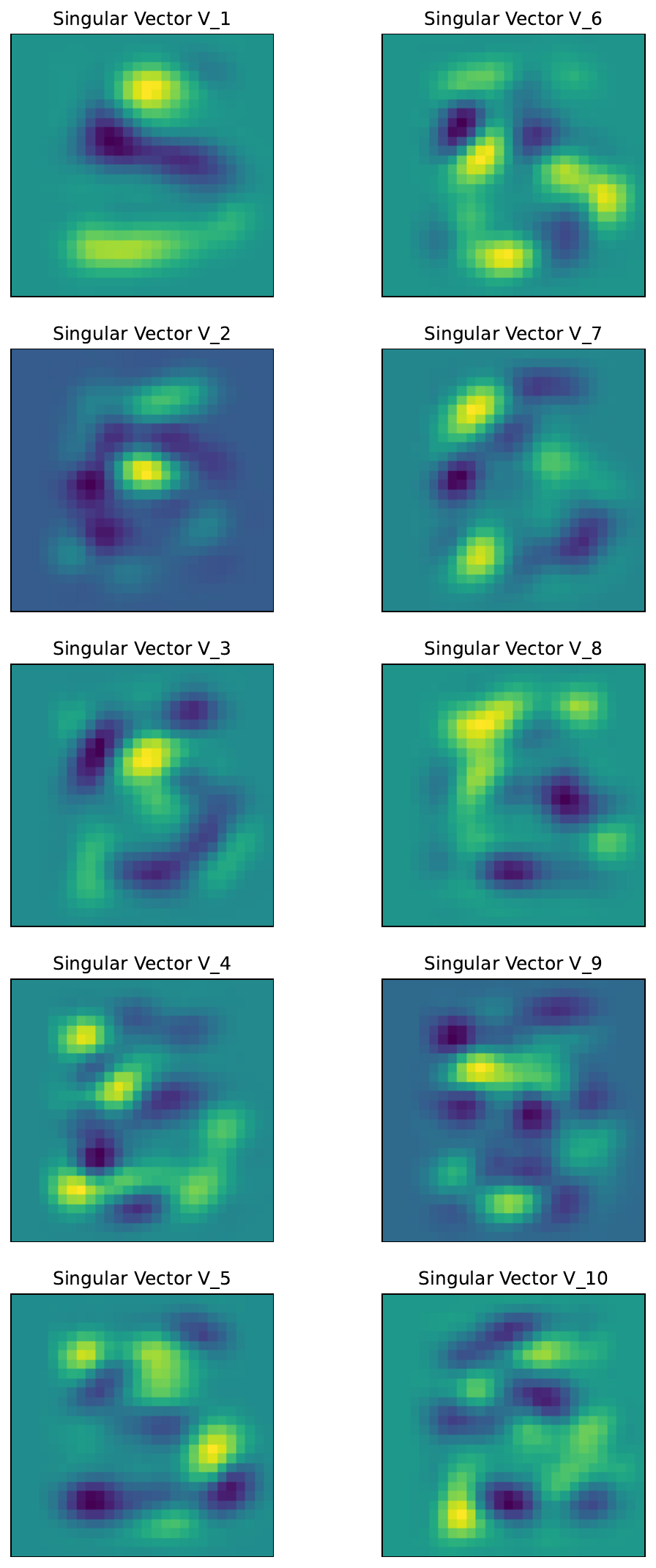}
    \caption{The right singular vectors with respect to the input image in Figure \ref{fig:projection_sam_15}.}
    \label{fig:right_vec_sam_15}
\end{figure}

\end{document}